\pdfoutput=1
\documentclass[11pt]{article}
\usepackage{acl}

\usepackage{times}
\usepackage{latexsym}
\usepackage{makecell}
\usepackage{multirow}
\usepackage{url}
\usepackage{adjustbox}
\usepackage{comment}

\usepackage[T1]{fontenc}

\usepackage[utf8]{inputenc}

\usepackage{microtype}

\title{Leveraging BERT Language Models for Multi-Lingual ESG Issue Identification}

\author{Elvys Linhares Pontes \and
  Mohamed Benjannet \\
  Trading Central Labs, Trading Central, Paris, France\\
  \texttt{\{elvys.linharespontes,mohamed.benjannet\}@tradingcentral.com} \\\AND
  Lam Kim Ming\\
  Trading Central, Hong Kong, China
}

\begin{document}
\maketitle
\begin{abstract}
Environmental, Social, and Governance (ESG) has been used as a metric to measure the negative impacts and enhance positive outcomes of companies in areas such as the environment, society, and governance. Recently, investors have increasingly recognized the significance of ESG criteria in their investment choices, leading businesses to integrate ESG principles into their operations and strategies.
The Multi-Lingual ESG Issue Identification (ML-ESG) shared task encompasses the classification of news documents into 35 distinct ESG issue labels. In this study, we explored multiple strategies harnessing BERT language models to achieve accurate classification of news documents across these labels. Our analysis revealed that the RoBERTa classifier emerged as one of the most successful approaches, securing the second-place position for the English test dataset, and sharing the fifth-place position for the French test dataset. Furthermore, our SVM-based binary model tailored for the Chinese language exhibited exceptional performance, earning the second-place rank on the test dataset.
\end{abstract}

\section{Introduction}

Financial markets and investors play a crucial role in advancing the transition towards a more sustainable economy by actively promoting investments in companies that adhere to ESG (Environment, Social, and Governance) principles\footnote{https://www.investopedia.com/terms/e/environmental-social-and-governance-esg-criteria.asp}. In today's landscape, there is a burgeoning interest among investors in assessing the sustainability performance of firms~\cite{su13073746}. Consequently, it becomes imperative to efficiently identify and extract pertinent information pertaining to companies' ESG strategies.

To facilitate this process, the application of NLP (Natural Language Processing) techniques tailored to the finance and ESG domain can significantly aid in the identification and processing of relevant information. By leveraging these advanced methods, valuable insights can be gleaned from vast amounts of financial data and reports, enabling informed investment decisions aligned with sustainable principles~\cite{armbrust2020computational,mehra2022esgbert}.

Indeed, \citet{armbrust2020computational} analyzed the impact of a company's environmental performance, on the connection between the company's disclosures and financial performance. The authors, discovered that the textual information in the Management’s Discussion and Analysis of Financial Conditions and Results of Operations section alone does not provide insights into the future financial performance of the company. However, they did find evidence that NLP methods can extract information about the environmental performance of the company.
\citet{mehra2022esgbert} focused on building a BERT-based model to predict two tasks: determining whether there was a change or no change in environmental scores; and identifying a positive or negative change (if any) in the environmental scores of companies based on ESG-related text found in their 10-Q filings. Their results demonstrated that their model can predict not only environmental risk scores but also assessing Social and Governance risk scores for companies.


The Multi-Lingual ESG Issue Identification (ML-ESG) shared task focus on the classification of ESG issue of news articles~\cite{chen2023ESG}.
Based on the MSCI ESG rating guidelines, the organizers created multilingual news articles and categorized them into 35 key ESG issues. The target languages include English, Chinese, and French, enhancing the task's cross-lingual scope and enriching the diversity of the dataset.

The main challenge of this task lies in accurately identifying the ESG issues discussed in each article. To address this challenge, the system must possess comprehensive knowledge about the specific ESG issues addressed in each article. In this study, we propose a range of strategies that leverage the capabilities of BERT language models. Among our various approaches, our RoBERTa classifier achieved outstanding results and securing the second-highest scores for the English test set, and sharing the fifth-place position for the French test dataset of the shared task. Additionally, our SVM-based binary model achieved the second-best results on the Chinese test dataset. These findings demonstrate the effectiveness of our proposed strategies in accurately classifying ESG issues in multilingual news articles.

\section{Multi-Lingual ESG Issue Identification shared task}
\label{sc:ml-esg}

The Multi-Lingual ESG Issue Identification (ML-ESG) shared task presents a compelling challenge focused on ESG issue identification. Drawing from the well-established MSCI ESG rating guidelines\footnote{https://www.msci.com/our-solutions/esg-investing/esg-ratings}, ESG-related news articles can be categorized into 35 distinct ESG key issues. For this task, participants are expected to devise systems capable of recognizing and classifying the specific ESG issue associated with an article~\cite{chen2023ESG}.

In essence, the objective of the ML-ESG shared task is to develop robust systems that demonstrate awareness of the ESG issues encompassed within each article. By accurately classifying the ESG issues, participants can effectively contribute to the advancement of ESG analysis and understanding within the domain of multi-lingual news articles.

\subsection{Datasets}

The organizers provided a multilingual datasets for Chinese, English and French languages. They annotated these datasets based on the  MSCI ESG rating guidelines. More precisely, these datasets are composed of news articles that were classified into 35 ESG key issues. The English and French datasets contain 1200 articles and the Chinese dataset contains 1000 articles. More details about the datasets are available at~\cite{chen2023ESG}.

\section{BERT-based approaches}
\label{sc:berts}

We applied several strategies to classify the article in the ESG issues classes. As BERT-based models has proved the performance of general and financial applications~\cite{9672027,linhares-pontes-etal-2022-using,yang2022esg}, our following strategies are based on the BERT models on their architectures.

\subsection{SVM+EE}

Inspired by the performance of semantic similarity~\cite{linhares-pontes-etal-2018-predicting} and the performance of the~\citet{linhares-pontes-etal-2022-using}'s model in classifying the ESG taxonomies, this approach analyzes the ESG issue classification by considering all articles pertaining to a specific ESG issue as similar, as they inherently share the same underlying semantic information. To facilitate this analysis, we employ the SBERT (Sentence-BERT) model~\cite{reimers-2019-sentence-bert}, which projects the articles onto a shared dimensional space.

To classify these paraphrased articles into their respective ESG issue classes, we employ a Support Vector Machine (SVM) model~\cite{PlattProbabilisticOutputs1999}. The SVM model is trained to analyze and categorize the articles based on their semantic similarity and the corresponding ESG issue classes.

In our methodology, we further enhance the classification process by incorporating the probability of each class provided by the SVM, along with the cosine distance between the SBERT representation of the article and the SBERT representation of the corresponding ESG issue definition (i.e. Esg issue Embeddings (EE)). This combined approach allows us to capture the semantic relationships between articles and ESG issue classes, enabling more accurate and robust classification results.

\subsection{RoBERTa}

We present an approach for article classification, leveraging the capabilities of RoBERTa-based language models~\cite{liu2019roberta} in conjunction with a feed-forward multi-layer perceptron. Our proposed RoBERTa classifier effectively captures contextual information within sentences, enabling accurate classification into distinct ESG issue classes.

To extract sentence context and facilitate classification, we utilize the representation of the special [CLS] token from the final layer of the BERT-based language models. Furthermore, we incorporate a feed-forward layer to enhance the classification process, accurately assigning input articles to their respective ESG issue classes.

\subsection{RoBERTa+EE}

Delving deeper into the realm of neural networks, we have extended the RoBERTa classifier by incorporating ESG issue embeddings (EE). More precisely, The architecture of our model integrates RoBERTa-based language models with article embeddings and ESG issue label definition embeddings using SBERT model. This integration enables us to perform a more comprehensive analysis of the article and classify it accurately into one of the ESG issue classes.

To extract the contextual information and aid in classification, we employ the representation of the special [CLS] token from the final layer of the BERT-based language models, along with the incorporation of article and ESG issues representations. By combining these representations, we capture a richer understanding of the article's content and its relationship to all ESG issues. Finally, our model incorporates a feed-forward layer that combines all this information on the classification of input articles into their respective ESG issue classes.

\subsection{RoBERTa+CNN+SVM}

This architecture leverages the combined strength of the RoBERTa language model, a Convolutional Neural Network (CNN)~\cite{OSheaN15}, and a SVM to extract diverse features from articles at various levels. Firstly, the RoBERTa language model generates token embeddings for an input article, taking into account its contextual information. Next, the CNN layer performs five convolutions on these token embeddings, capturing different features within the contextualized tokens. This enables the CNN to extract local patterns and features from the textual data, effectively capturing important information across different scales. The final layer of the neural network consists of a feed-forward layer that classifies the output of the CNN into respective ESG issue classes.

Once the neural network model is trained, we use an SVM model to classify the articles into ESG issue classes. To accomplish this, we feed the representation of the articles, which is generated by the previously described CNN, as input to the SVM model.

\section{Experimental setup and evaluation}
\label{sc:experimental}

\subsection{Evaluation metrics}

All system outputs were evaluated by examining key performance metrics such as precision, recall, and F1-score. Precision represents the number of well predicted positives divided by all the positives predicted. Recall measures the number of well predicted positives divided by the total number of positives. Finally, the F1-score takes into account both precision and recall, providing a balanced assessment of the system's performance in identifying and classifying ESG issues.

\subsection{Training procedure}

The dataset provided by the organizers was divided into two parts: 70\% was allocated for training purposes, while the remaining 30\% was set aside for development.
To set up the meta-parameters for each approach, we used the development dataset. Our SBERT model uses the pre-trained model `\textit{sentence-transformers/paraphrase-multilingual-mpnet-base-v2}'\footnote{https://huggingface.co/sentence-transformers/paraphrase-multilingual-mpnet-base-v2} to generate the embeddings of articles and ESG issue definitions for all languages in the same dimensional space. 

For all approaches, we created two models versions with different training datasets. In the first version, our classifier model was trained exclusively on target language data (monolingual). For the second version, we combined English and French training data (multilingual) to train our classifier models.
The SVM models were trained using a linear kernel to classify the article embeddings provided by the SBERT model into ESG issue labels.
For BERT-based models, the last layer incorporates a dropout of 0.2 to improve the model's generalization ability. Additionally, we used the \textit{'xlm-roberta-large'} for the multilingual training and French and Chinese models, and \textit{'roberta-large'} for the English model.

Once the meta-parameters were defined, we proceeded to train the model using both the train and development datasets.

\subsection{Experimental evaluation}

In order to select the best models for the ML-ESG shared task, we evaluated all models on the development dataset.
For the English dataset, the RoBERTa classifier using only monolingual data achieved the best results. The use of SBERT to represent the article and ESG issue embeddings did not add relevant information to improve the performance of our classifiers. Interestingly, despite the simplest model being the SVM+EE trained on the English dataset, it achieved similar results to the BERT+CNN+SVM model.

\begin{table}[h]
\centering
\begin{tabular}{cccc}
\hline
\textbf{Approach} & \textbf{Acc.} & \textbf{MF1} & \textbf{WF1} \\\hline
\makecell{SVM+EE\\monolingual} & 0.66              & 0.59                       & 0.65                          \\
\makecell{SVM+EE\\multilingual} & 0.61              & 0.57                       & 0.6                           \\
\textbf{\makecell{RoBERTa\\monolingual}} & \textbf{0.71}              & \textbf{0.67}                       & \textbf{0.71}                          \\
\makecell{RoBERTa\\multilingual} & 0.69              & 0.67                       & 0.69                          \\
\makecell{RoBERTa+EE\\monolingual}  & 0.7               & 0.63                       & 0.69                          \\
\makecell{RoBERTa+EE\\multilingual}  & 0.69               & 0.68                       & 0.69                          \\
\makecell{BERT+CNN+SVM\\multilingual} & 0.66              & 0.61                         & 0.65                  \\\hline
\end{tabular}
\caption{English results for the development dataset. The best results are highlighted in bold. Acc: accuracy, MF1: macro average f-score, and WF1: weighted averaged f-score.}
\end{table}

Differently from the English model, the use of multilingual data to train our models improve the results when compared with their respective monolingual version. The RoBERTa classifier using multilingual data achieved the best results.

\begin{table}[h]
\centering
\begin{tabular}{cccc}
\hline
\textbf{Approach}                & \textbf{Acc.} & \textbf{MF1} & \textbf{WF1} \\\hline
\makecell{SVM+EE\\monolingual} & 0.66              & 0.63               & 0.66                  \\
\makecell{SVM+EE\\multilingual} & 0.69              & 0.69               & 0.68                  \\
\makecell{RoBERTa\\monolingual}  & 0.71              & 0.7                & 0.71                  \\
\textbf{\makecell{RoBERTa\\multilingual}} & \textbf{0.73}              & \textbf{0.72}               & \textbf{0.73}                  \\
\makecell{RoBERTa+EE\\monolingual}   & 0.72              & 0.72               & 0.72                  \\
\makecell{RoBERTa+EE\\multilingual}  & 0.73              & 0.71               & 0.72                  \\\hline
\end{tabular}
\caption{French results for the development dataset. The best results are highlighted in bold. Acc: accuracy, MF1: macro average f-score, and WF1: weighted averaged f-score.}
\end{table}

For the Chinese model, we utilized the SVM model trained on the representation provided by the SBERT model. This allowed us to classify the ESG issue classes in a binary mode. We employed a binary classifier for each ESG issue and then selected the ESG issue classes with the highest probabilities as the output of our classifiers.

\subsection{Official results}

The organizers published the official results for each language. Our models were labeled as TradingCentralLabs (TCL). For the Chinese dataset, we submitted three runs (Table~\ref{tb:official-chinese}). All runs use the same model but the number of ESG issue labels output change for each one of them. More precisely, the run 1 provides only the most probable ESG issue class as answer, the run 2 uses the top 2 most probable classes and run 3 uses the top 3 ESG issue classes.

The superior performance of run 3, compared to other runs, can be attributed to the prevalence of multiple ESG issue classes for each article in the gold data. It is worth noting that many examples in the gold data encompassed multiple classes, with some cases containing up to 8 classes. As the Chinese test data consisted of several examples with multiple ESG issue classes, run 3 achieved the best results by predicting the top 3 ESG issue classes.

This ability to accommodate the presence of multiple classes in certain cases elucidates why run 3 outperformed the others. Finally, our run 3 secured the second position in the official ranking for the Chinese data.

\begin{table}[h]
\centering
\begin{tabular}{cccc}
\hline
\textbf{Runs} & \textbf{Mic. F1} & \textbf{Mac. F1} & \textbf{WF1} \\\hline
\textbf{CheryFS\_2}            & \textbf{0.3914}   & \textbf{0.1799}   & \textbf{0.3921}      \\
\textit{TCL\_3} & \textit{0.2790}   & \textit{0.1367}   & \textit{0.2633}      \\
TCL\_2 & 0.2665   & 0.1032   & 0.2332      \\
TCL\_1 & 0.2115   & 0.0730   & 0.1791      \\\hline
\end{tabular}
\caption{Official results for the Chinese test data. The best results are highlighted in bold and our best results are in italic. Micro (Mic.), macro (Mac.) and weighted (WF1) F1-score.}
\label{tb:official-chinese}
\end{table}

For the English runs, we employed three different approaches (Table~\ref{tb:official-english}). The run 1 corresponds to the RoBERTa classifier (monolingual), the run 2 corresponds to the RoBERTa+EE classifier (monolingual) and the run 3 employed the BERT+CNN+SVM classifier (multilingual). As expected, the run 1 outperformed the other runs by generalizing much better ESG issue labels. While the gap in the results on the development data was quite small, the run 1 increased the gap compared to the runs 2 and 3 on test data. The RoBERTa classifier obtained the second-place ranking, achieving a score just 2 points lower than the NCMU\_1 model.  

\begin{table}[h]
\centering
\begin{tabular}{cccc}
\hline
\textbf{Runs} & \textbf{P} & \textbf{R} & \textbf{F1} \\\hline
\textbf{NCMU\_1}                       & \textbf{0.69}      & \textbf{0.70}   & \textbf{0.69}     \\
\textit{TradingCentralLabs\_1}         & \textit{0.67}      & \textit{0.68}   & \textit{0.67}     \\
TradingCentralLabs\_2         & 0.61      & 0.63   & 0.61     \\
TradingCentralLabs\_3         & 0.59      & 0.59   & 0.58     \\\hline
\end{tabular}
\caption{Official results for the English test data. The best results are highlighted in bold and our best results are in italic. P: precision, R: recall, and F1: f1-score.}
\label{tb:official-english}
\end{table}

We submitted three runs for the French test data as well (Table~\ref{tb:official-french}). Run 1 used the SVM+EE model, run 2 employed the RoBERTa classifier, and run 3 utilized RoBERTa+EE. All three runs were trained using both the English and French training datasets.

Our run 2 achieved a commendable fifth place in terms of F1-score, falling just 3 points short of the top position. This highlights its competitive performance and showcases its potential for accurately classifying ESG issue classes in the French language.

\begin{table}[h]
\centering
\begin{tabular}{cccc}
\hline
\textbf{Runs} & \textbf{P} & \textbf{R} & \textbf{F1} \\\hline
\textbf{Jetsons\_2}                         & \textbf{0.80}      & \textbf{0.79}   & \textbf{0.78}         \\
\textit{TradingCentralLabs\_2}              & \textit{0.76}      & \textit{0.76}   & \textit{0.75}         \\
TradingCentralLabs\_3              & 0.74      & 0.74   & 0.73         \\
TradingCentralLabs\_1              & 0.73      & 0.72   & 0.71         \\\hline
\end{tabular}
\caption{Official results for the French test data. The best results are highlighted in bold and our best results are in italic. P: precision, R: recall, and F1: f1-score.}
\label{tb:official-french}
\end{table}

\section{Conclusion}
\label{sc:conclusion}

This paper presents the participation of Trading Central Labs in the Multi-Lingual ESG Issue Identification evaluation campaign for financial documents. Our objective was to accurately classify financial documents into ESG issue labels, and to achieve this, we proposed several BERT-based models.

Among our models, the one based on the RoBERTa classifier emerged as a standout performer, securing the second-place ranking for the English language. It was just 2 points behind the top-performing model. Additionally, our RoBERTa-based model also demonstrated its capability in the French language, sharing the fifth place, with a slight margin of 3 points from the leading result. Finally, our SVM-based model for the Chinese data claimed the second-place ranking, further illustrating our competitive performance.

These results underscore the competitive edge and potential of our models in accurately classifying ESG issue classes across different languages.

\section*{Acknowledgements}
This work has been partially supported by the France Relance project (grant agreement number ANR-21-PRRD-0010-01).

\bibliography{anthology}

\end{document}